\documentclass[a4paper,fleqn]{cas-sc}

\usepackage[authoryear,longnamesfirst]{natbib}

\usepackage{algpseudocode}
\usepackage{longtable}

\usepackage[ruled]{algorithm2e}
\usepackage{amsmath,amsthm,amssymb,hyperref}

\def\tsc#1{\csdef{#1}{\textsc{\lowercase{#1}}\xspace}}
\tsc{WGM}
\tsc{QE}
\tsc{EP}
\tsc{PMS}
\tsc{BEC}
\tsc{DE}

\usepackage[utf8]{inputenc}
\usepackage{bm}

\theoremstyle{property}
\usepackage{enumitem}
\usepackage{multirow}
\usepackage{adjustbox}
\usepackage{makecell}

\begin{document}

\let\WriteBookmarks\relax
\def\floatpagepagefraction{1}
\def\textpagefraction{.001}

\shorttitle{Training Dynamics of a 1.7B LLaMa Model: A Data-Efficient Approach}
\shortauthors{Li et~al.}

\title [mode = title]{Training Dynamics of a 1.7B LLaMa Model: A Data-Efficient Approach}

\author[1,3]{Miles Q. Li}[orcid=0000-0001-7091-3268]
\ead{miles.qi.li@mail.mcgill.ca}

\author[1]{Benjamin C. M. Fung}[orcid=0000-0001-8423-2906]\cormark[1]
\ead{ben.fung@mcgill.ca}

\author[2]{Shih-Chia Huang}[orcid=0000-0002-6896-3415]\cormark[1]
\ead{schuang@ntut.edu.tw}

\address[1]{School of Information Studies, McGill University, Montreal, Canada}

\address[2]{Department of Electronic Engineering, National Taipei University of Technology, Taipei, Taiwan}

\address[3]{InfiniteOptimization AI Lab, Montreal, Canada}

\cortext[cor1]{Corresponding author.}

\begin{abstract}
Pre-training large language models is a complex endeavor influenced by multiple factors, including model architecture, data quality, training continuity, and hardware constraints. In this paper, we share insights gained from the experience of training DMaS-LLaMa-Lite, a fully open-source, 1.7-billion-parameter, LLaMa-based model, on approximately 20 billion tokens of carefully curated data. We chronicle the full training trajectory, documenting how evolving validation loss levels and downstream benchmarks reflect transitions from incoherent text to fluent, contextually grounded output. Beyond pre-training, we extend our analysis to include a post-training phase focused on instruction tuning, where the model was refined to produce more contextually appropriate, user-aligned responses. We highlight practical considerations such as the importance of restoring optimizer states when resuming from checkpoints, and the impact of hardware changes on training stability and throughput. While qualitative evaluation provides an intuitive understanding of model improvements, our analysis extends to various performance benchmarks, demonstrating how high-quality data and thoughtful scaling enable competitive results with significantly fewer training tokens. By detailing these experiences and offering training logs, checkpoints, and sample outputs, we aim to guide future researchers and practitioners in refining their pre-training strategies. The training script is available on Github and the model checkpoints are available on Huggingface. 
\end{abstract}


\maketitle

\section{Introduction}

Large language models (LLMs) have achieved unprecedented performance across a variety of natural language understanding and generation tasks. These advancements have been driven by innovations in model architectures, pre-training methodologies, and the availability of large-scale, high-quality data \citep{brown2020language, chowdhery2022palm, touvron2023llama}. However, despite significant progress, many aspects of LLM training remain underexplored. Practical challenges—ranging from data curation and training stability to qualitative and quantitative evaluation—play a critical role in determining model performance but are often overlooked in favor of final benchmark results.

In this paper, we present our experience pre-training and post-training \textbf{DMaS-LLaMa-Lite}, a 1.7-billion-parameter LLaMa-based model, on 20 billion tokens of carefully curated data. Unlike many existing studies that focus solely on the final outputs of large-scale training, we emphasize the training process itself and the insights gleaned from it. Specifically, we document the following key aspects of our work:

\begin{enumerate}
\item \textbf{Training Dynamics:} We analyze how validation loss and downstream benchmarks (e.g., Hellaswag, ARC) evolve over 40,000+ training steps and correlate these metrics with improvements in text fluency, coherence, and factual accuracy.

\item \textbf{Practical Lessons Learned:} We demonstrate the critical importance of restoring optimizer states when resuming training from checkpoints, as failure to do so results in abrupt loss spikes and degraded model performance. We also explore the effects of hardware transitions (e.g., switching between single-GPU and multi-GPU setups) on training stability and efficiency.

\item \textbf{The Impact of High-Quality Data:} By leveraging a carefully curated subset of the FineWeb-Edu dataset, we show that high-quality training data enables strong model performance, even when trained on significantly fewer tokens compared to other models such as TinyLLaMa.

\item \textbf{Qualitative Observations:} Using diverse evaluation prompts, we track how the model transitions from repetitive, semantically incoherent outputs to fluent, contextually appropriate completions. These observations highlight how different stages of training contribute to linguistic and factual improvements.

\item \textbf{Post-Training Instruction Tuning:} We refine the pretrained model using parameter-efficient fine-tuning, leveraging high-quality instruction datasets. We documented the model's behavioral change from the process and the model's sensitivity to the prompt.
\end{enumerate}

Our findings collectively offer insights into the practical challenges and trade-offs encountered during the pre-training of LLMs. By sharing detailed training logs, checkpoint artifacts, and sample completions, we aim to support reproducibility and provide guidance for both researchers and practitioners seeking to optimize their own training strategies. Importantly, this work serves as a foundation for future studies that may uncover additional best practices, methodological refinements, or novel evaluation techniques in the pre-training of large language models. The training script is available on Github\footnote{\url{https://github.com/McGill-DMaS/DMaS-LLaMa-Lite-Training-Code}}. The model checkpoints are available on Huggingface\footnote{\url{https://huggingface.co/collections/McGill-DMaS/dmas-llama-lite-6761d97ba903f82341954ceb}}.

The remainder of this paper is organized as follows: Section ~\ref{ModelConfiguration} describes the model architecture, training setup, and data preprocessing. Section ~\ref{pretrain} presents our findings from the pre-training experience, including quantitative and qualitative evaluations, and a discussion of broader implications. Section ~\ref{posttrain} details the post-training phase, including instruction tuning and qualitative analysis. Finally, Section ~\ref{conclu} concludes with key takeaways and directions for future work.

\section{Model Configuration and Training Setup}
\label{ModelConfiguration}

\subsection{Model Architecture}
Our model, DMaS-LLaMa-Lite, is a 1.7B-parameter (1716M) variant of LLaMa \citep{touvron2023llama} adapted to use a GPT-2 tokenizer. The configuration details are presented in Table~\ref{tab:model_config}. The model features 36 transformer layers, each with a hidden size of 2048 and an intermediate size of 5120 in the feed-forward layers. We employ 32 attention heads, grouped into 8 key-value heads, and use a rotary positional embedding scheme. Norm layers use RMS normalization \citep{zhang2019root}, and the model’s non-linear activation is SiLU \citep{hendrycks2016gaussian}. The model is trained on tokens produced by the GPT-2 tokenizer.

\begin{table}[!h]
\centering
\caption{Model configuration for the 1.7B-parameter DMaS-LLaMa-Lite model.}
\begin{tabular}{l|l}
\toprule
\textbf{Attribute}          & \textbf{Value}             \\ 
\midrule
Architectures               & LlamaForCausalLM          \\ \hline
Attention Bias              & False                     \\ \hline
Attention Dropout           & 0.0                       \\ \hline
Hidden Activation           & SiLU                      \\ \hline
Hidden Size                 & 2048                      \\ \hline
Intermediate Size           & 5120                      \\ \hline
Max Position Embeddings     & 1024                      \\ \hline
MLP Bias                    & False                     \\ \hline
Number of Attention Heads   & 32                        \\ \hline
Number of Hidden Layers     & 36                        \\ \hline
Number of Key-Value Heads   & 8                         \\ \hline
RMS Norm Epsilon            & $1 \times 10^{-5}$        \\ \hline
ROPE Scaling                & Null                      \\ \hline
ROPE Theta                  & 500000                    \\ \hline
Vocabulary Size             & 50257                     \\
\bottomrule
\end{tabular}
\label{tab:model_config}
\end{table}

\subsection{Training Data and Preprocessing}
The training was conducted using a subset of the \texttt{HuggingFaceFW/fineweb-edu 100BT}~\citep{penedo2024fineweb} dataset, derived from the broader FineWeb dataset. FineWeb-Edu comprises approximately 1.3 trillion tokens of high-quality educational content, curated through a meticulous multi-stage filtering process. FineWeb itself is a 15-trillion-token dataset sourced from 96 Common Crawl snapshots, designed to improve LLM performance through advanced filtering and deduplication methods.

FineWeb-Edu emphasizes educational text, selected using a classifier trained on synthetic annotations generated by Llama-3-70B-Instruct~\citep{dubey2024llama}. This classifier was fine-tuned to prioritize grade-school and middle-school-level knowledge while excluding overly technical content. The dataset employs custom filters targeting document structure and content quality, ensuring a high proportion of semantically coherent and educationally relevant text.

Deduplication was performed on a per-snapshot basis using MinHash clustering, which identifies and removes highly similar documents, thus preventing redundancy while preserving dataset diversity. Additional heuristic filters were applied to refine text quality, such as removing documents with low punctuation density or high proportions of repeated lines.

The use of FineWeb-Edu, with its emphasis on quality and diversity, contributed to the model's ability to perform well on text completion tasks, supporting efficient training with a reduced token count.

\subsection{Optimizer and Learning Rate Schedule}
We employ the AdamW optimizer \citep{loshchilov2017decoupled} with an initial learning rate of $6 \times 10^{-4}$. The learning rate is linearly warmed up for a short initial phase (e.g., $\sim$X steps; exact value omitted here) and then decayed following a schedule that we will report in the final version of the paper. Gradient clipping is applied with a maximum norm threshold to maintain training stability.

\subsection{Training Procedure and Checkpoints}
The model is trained for more than 40,000 steps, each step consuming 0.5 million tokens, resulting in a substantial fraction of the 100-billion-token corpus being processed. Checkpoints are saved every 100 steps, allowing for a fine-grained inspection of the training trajectory. This frequent checkpointing also enables systematic evaluation of intermediate models across a range of cross-entropy validation losses.

The training was conducted on one or two RTX A6000 GPUs, reflecting realistic constraints in research and production environments. When using two GPUs, PyTorch's DistributedDataParallel (DDP)~\citep{paszke2019pytorch} was employed to ensure efficient data and model parallelism. Transitions in the hardware setup (e.g., moving from a single RTX A6000 GPU to dual RTX A6000 GPUs) served as natural breakpoints, allowing us to examine how changes in training continuity, combined with optimizer state restoration or omission, affected validation loss and model quality.

\section{Pre-training}
\label{pretrain}

\subsection{Validation Loss Trajectories and Training Metrics}

\begin{figure}[htbp]
\centering
\includegraphics[width=16cm]{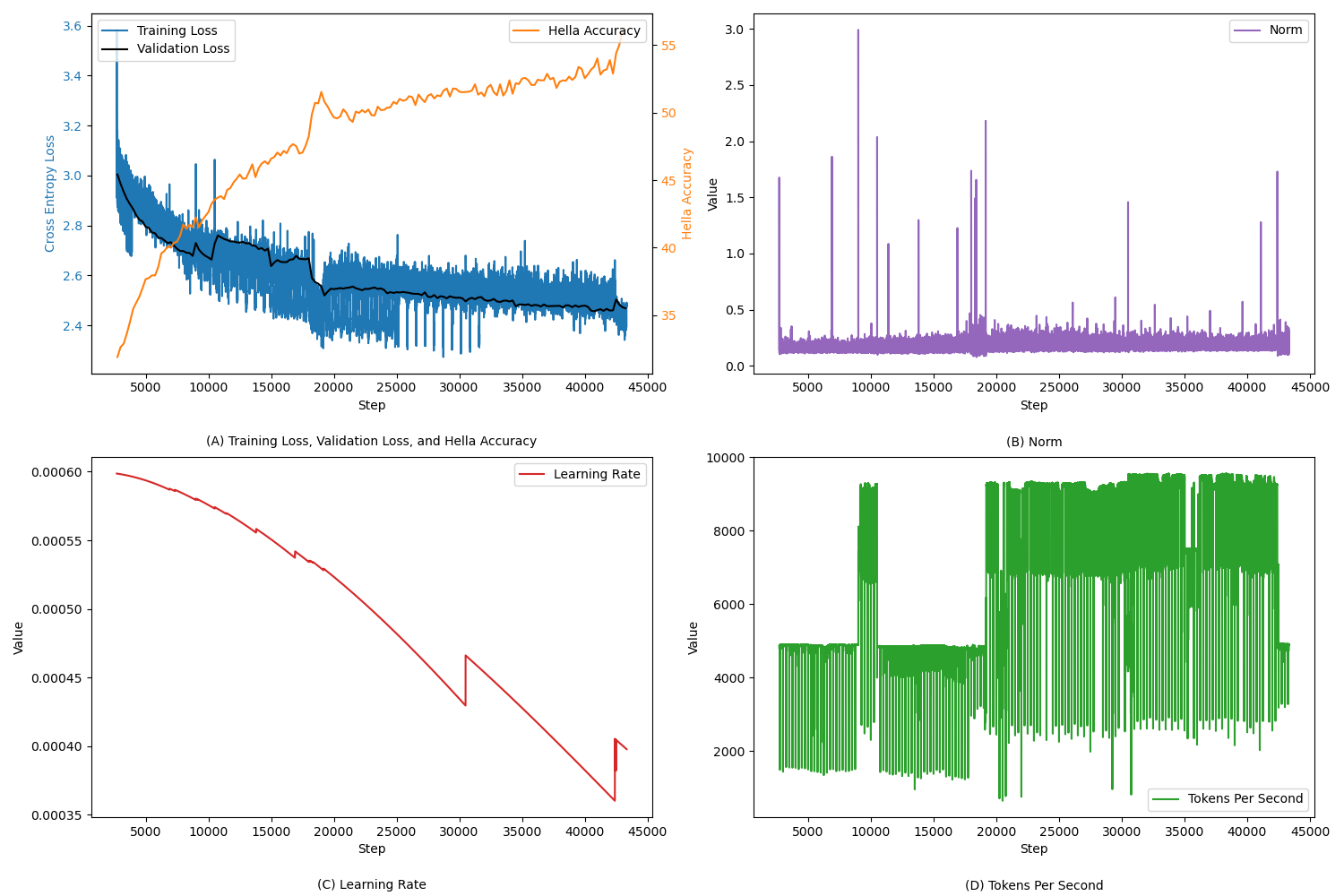}
\caption{Training logs visualizing training and validation loss, Hella accuracy, learning rate decay, norm behavior, and tokens processed per second over the course of 40,000+ steps.}
\label{fig:training_logs}
\end{figure}

Figure~\ref{fig:training_logs} provides a comprehensive visualization of the training process, capturing critical metrics such as training and validation loss, Hella accuracy, learning rate, gradient norm, and tokens processed per second. This visualization supplements the validation loss curve with additional context about the dynamics of training.

As shown in Figure~\ref{fig:training_logs}, validation loss decreases steadily over the course of 40,000+ steps. The upward trajectory of Hella accuracy mirrors the decline in validation loss, providing an additional indicator of the model’s qualitative improvements. Hella accuracy reflects the alignment of generated outputs with task-specific correctness criteria, reinforcing that lower validation loss corresponds to better adherence to instructions and prompts.

A notable finding is the impact of resuming training without restoring optimizer states. This practice led to temporary spikes of validation loss, Hellaswag accuracy, gradient norm and degraded performance. For example, Figure~\ref{fig:training_logs} (A) shows the difference in validation loss trajectories before and after training restart at step 8750 and step 10250. From Figure~\ref{fig:training_logs} (D) we can tell that the reason was that we switched between training on one and two GPUs, resuming training without optimizer states resulted in a increase in validation loss, requiring several thousand steps to recover.

\textbf{Key Takeaways:}
The combined view of validation loss, Hella accuracy, and other metrics reveals a strong correlation between decreasing loss and qualitative improvements in model outputs. Furthermore, the importance of maintaining training continuity is reinforced by the observed loss spikes and reduced performance when optimizer states are not restored during checkpoint-based restarts.

\subsection{Qualitative Evaluation of Generated Text}

We evaluated the model outputs at various checkpoints using a set of prompts designed to test several criteria:
\begin{itemize}
    \item \textbf{Coherence (C):} Does the completion logically follow from the given prompt? Are the ideas well-connected and form a consistent narrative or argument?
    \item \textbf{Fluency and Grammar (F):} Is the text grammatically correct and stylistically natural? Are there spelling or punctuation errors?
    \item \textbf{Relevance to Prompt (R):} Does the completion address the topic or instruction provided in the prompt? Does it stay on-point rather than drifting into unrelated content?
    \item \textbf{Factual Accuracy (FA):} Are factual claims correct? Does the model misstate well-known facts or historical information?
    \item \textbf{Depth and Completeness (D):} Does the completion provide a thorough and informative response given the complexity of the prompt?
    \item \textbf{Creativity (Cr):} For narrative or imaginative prompts, does the model produce original, interesting, or vivid content?
\end{itemize}

\begin{longtable}{c|p{14cm}}
\caption{Generated completions at different training steps.} \label{tab:battle_of_hastings} \\
\toprule
Checkpoint & Generated Text \\
\midrule
\endfirsthead

\toprule
Checkpoint & Generated Text \\
\midrule
\endhead

\midrule
\multicolumn{2}{r}{{Continued on next page}} \\
\midrule
\endfoot

\bottomrule
\endlastfoot
Prompt & "The Pyramids of Giza in Egypt are some of the oldest man-made structures in the world. Visitors often note"\\
\midrule
2700 & "The Pyramids of Giza in Egypt are some of the oldest man-made structures in the world. Visitors often note the impressive size and precision of the structures, marveling at how such feats of engineering were accomplished thousands of years ago without modern tools." \\
\hline
3300 & "The Pyramids of Giza in Egypt are some of the oldest man-made structures in the world. Visitors often note the mystery surrounding their construction, as well as the stunning alignment with celestial bodies." \\
\hline
5100 & "The Pyramids of Giza in Egypt are some of the oldest man-made structures in the world. Visitors often note the grandeur and historical significance of these monuments, which continue to captivate people from around the globe." \\
\hline
15700 & "The Pyramids of Giza in Egypt are some of the oldest man-made structures in the world. Visitors often note their intricate design, which reflects the advanced knowledge of ancient Egyptian engineers and architects." \\
\hline
35000 & "The Pyramids of Giza in Egypt are some of the oldest man-made structures in the world. Visitors often note the awe-inspiring scale and the enduring mystery of how they were constructed with such precision." \\
\hline
43250 & "The Pyramids of Giza in Egypt are some of the oldest man-made structures in the world. Visitors often note the breathtaking beauty and the remarkable testament to the ingenuity of ancient civilizations." \\
\midrule
Prompt & "The Mona Lisa, painted by Leonardo da Vinci, is one of the most famous artworks in the world. It is notable for"\\
\midrule
2700 & "The Mona Lisa, painted by Leonardo da Vinci, is one of the most famous artworks in the world. It is notable for its enigmatic smile and the use of sfumato technique to create a lifelike expression." \\
\hline
3300 & "The Mona Lisa, painted by Leonardo da Vinci, is one of the most famous artworks in the world. It is notable for its detailed background, which demonstrates da Vinci's mastery of perspective." \\
\hline
5100 & "The Mona Lisa, painted by Leonardo da Vinci, is one of the most famous artworks in the world. It is notable for its captivating expression, which has intrigued viewers and art historians for centuries." \\
\hline
15700 & "The Mona Lisa, painted by Leonardo da Vinci, is one of the most famous artworks in the world. It is notable for the subtle play of light and shadow on the subject's face, a hallmark of Renaissance art." \\
\hline
35000 & "The Mona Lisa, painted by Leonardo da Vinci, is one of the most famous artworks in the world. It is notable for its mysterious allure and the innovative techniques da Vinci employed to bring the portrait to life." \\
\hline
43250 & "The Mona Lisa, painted by Leonardo da Vinci, is one of the most famous artworks in the world. It is notable for its timeless beauty and the profound influence it has had on art and culture throughout history." \\

\midrule
Prompt & "The Battle of Hastings, a key event in English history, occurred in the year 1066. It resulted in significant changes to the ruling class of England. Following this battle"\\
\midrule
2700 & "The Battle of Hastings, a key event in English history, occurred in the year 1066. It resulted in significant changes to the ruling class of England. Following this battle, King Henry VIII was crowned king of England on January 1st 1485 at Westminster Abbey. King John's War (1315-14)." \\
\hline
3300 & "The Battle of Hastings, a key event in English history, occurred in the year 1066. It resulted in significant changes to the ruling class of England. Following this battle, the Normans were able to establish their own rule over England and established a powerful military force that would be able to defeat any other army in Europe at the time." \\
\hline
5100 & "The Battle of Hastings, a key event in English history, occurred in the year 1066. It resulted in significant changes to the ruling class of England. Following this battle, King Harold II was crowned as king by his Norman supporters and became known as 'the Conqueror'." \\
\hline
15700 & "The Battle of Hastings, a key event in English history, occurred in the year 1066. It resulted in significant changes to the ruling class of England. Following this battle was an era where the Normans were able to conquer and control much of Europe including France, Spain, Italy, Germany, Poland, Hungary, etc." \\
\hline
35000 & "The Battle of Hastings, a key event in English history, occurred in the year 1066. It resulted in significant changes to the ruling class of England. Following this battle, William became King and his descendants would rule over Britain for many years until Henry VIII broke away from Rome’s authority by declaring himself Head of the Church (the Pope)." \\
\hline
43250 & "The Battle of Hastings, a key event in English history, occurred in the year 1066. It resulted in significant changes to the ruling class of England. Following this battle, William I became King and he was known as William The Conqueror (1066-1087)." \\
\bottomrule
\end{longtable}

Table~\ref{tab:evaluation_scores} summarizes the averaged scores across the evaluated criteria for each checkpoint. Initially, the model receives middling marks on attributes such as Coherence and Fluency, reflecting a nascent grasp of narrative construction and linguistic form. Over time, as training steps increase, the gradual upward trend in categories like Depth and Completeness suggests that the model becomes not only more reliable at presenting known information but also more adept at elaborating thoughtfully on prompts. Nevertheless, the Factual Accuracy (FA) score remains more modest, with even the most extensively trained checkpoint (43250) receiving only a 5 in this category. This relatively muted improvement in factual accuracy underscores the inherent difficulty of producing historically and contextually precise content, even at later stages of model refinement. It indicates that the model is still not sufficiently trained in factual domains, reinforcing observations that large language models often require extensive training—and targeted data coverage—to reduce hallucinations. This also helps explain why earlier-generation models (such as ChatGPT-3.5 or LLaMa 1~\citep{touvron2023llama} and 2~\citep{touvron2023llama2}), trained with fewer tokens, were more prone to factual distortions, whereas newer models (ChatGPT-4~\citep{openai2023gpt4}, LLaMa 3~\citep{dubey2024llama} and Qwen 2.5~\citep{yang2024qwen2}) with more comprehensive training corpora tend to display far fewer factual errors. In essence, these aggregated scores provide a quantitative reflection of a well-known progression in language model development: as training deepens and data coverage broadens, models transition from producing disjointed or factually dubious content toward more coherent, contextually appropriate, and accurate outputs—though even the most advanced checkpoints still leave room for improvement in factual domains.

Table~\ref{tab:battle_of_hastings} presents completions for three prompt examples. The first prompt, \textit{"The Pyramids of Giza in Egypt are some of the oldest man-made structures in the world. Visitors often note"}, is relatively straightforward and factually well-known, focusing on describing a widely recognized historical landmark. The second prompt, \textit{"The Mona Lisa, painted by Leonardo da Vinci, is one of the most famous artworks in the world. It is notable for"}, also deals with a familiar cultural artifact whose key attributes (e.g., enigmatic smile, Renaissance painting techniques) are commonly referenced. In contrast, the third prompt, \textit{"The Battle of Hastings, a key event in English history, occurred in the year 1066. It resulted in significant changes to the ruling class of England. Following this battle"}, involves historical specificity and requires more precise factual recall.

For the Pyramids of Giza and Mona Lisa prompts, even early checkpoints produced relatively coherent and factually aligned completions. At checkpoint 2700, the generated text for the Pyramids of Giza prompt accurately describes their “impressive size and precision,” a well-established fact often cited in historical and travel literature. Similarly, the completion for the Mona Lisa at the same checkpoint references its “enigmatic smile and the use of sfumato,” demonstrating the model’s ability to align with canonical art historical commentary. These early-stage outputs suggest that the model may have more readily internalized broadly known cultural and historical touchpoints, resulting in fewer factual distortions for prompts centered on widely recognized content.

As training progresses, the completions for these two prompts become more polished, with improved stylistic fluency and contextual depth. While the essence of the content remains consistent—emphasizing mystery, engineering feats, aesthetic qualities, and cultural significance—the later checkpoints (e.g., 35000 and 43250) refine these attributes. For the Mona Lisa, checkpoint 43250 highlights its “timeless beauty” and “profound influence,” offering a more thematically rich and contextually appropriate summary. Similarly, by the later checkpoints, the descriptions of the Pyramids incorporate nuanced phrases like “enduring mystery” or reflect “advanced knowledge” of ancient engineers, demonstrating an increasingly sophisticated grasp of the subject matter.

In contrast, the more historically complex Battle of Hastings prompt initially posed greater challenges. For early checkpoints (e.g., checkpoint 2700, checkpoint 3300), the generated text is often incoherent and factually incorrect. For instance, checkpoint 2700 introduces King Henry VIII (born centuries after the Battle of Hastings) as the crowned king. Such completions highlight the model’s early inability to handle historical contexts accurately.

For intermediate checkpoints (e.g., checkpoint 5100, checkpoint 15700), the outputs improve in coherence, but factual inaccuracies persist. For example, checkpoint 5100 incorrectly states that Harold II, who died in the battle, was crowned as king. At checkpoint 15700, the model overgeneralizes, attributing exaggerated territorial control to the Normans.

By checkpoint 35000, the text becomes significantly more coherent and contextually relevant, though it introduces minor inaccuracies, such as the association with Henry VIII. At checkpoint 43250, the output demonstrates marked improvement, correctly identifying William I as "William The Conqueror" and providing accurate dates and titles.

Overall, these results suggest that while the model can produce coherent, factually aligned completions for widely known cultural and historical subjects from relatively early training stages, it requires more extensive training to handle historically specific prompts with precision. This pattern highlights the interplay between general cultural knowledge, which may be more readily accessible to the model, and the acquisition of precise historical facts, which appear to be more challenging and demand longer training or more targeted historical representations.

These observations align with the scores shown in Table~\ref{tab:evaluation_scores}: less sufficiently trained language models are more prone to “hallucination”—the production of factually incorrect or contextually inconsistent details—while more extensively trained models show a marked reduction in such errors. As the model’s training proceeds, it appears to acquire a stronger grounding in historical context, produce more thematically coherent narratives, and adhere more closely to factual accuracy. In other words, increased training not only enhances fluency and coherence but also mitigates the tendency to fabricate details, resulting in outputs that are both stylistically refined and factually sound.

This example underscores the relationship between decreasing validation loss and qualitative improvements in text generation. As the model trains, it transitions from producing semantically incoherent text to generating accurate and contextually relevant outputs. However, occasional factual inaccuracies even at advanced checkpoints highlight limitations inherent in the current training approach. These findings suggest that evaluating the model using prompts rich in historical, scientific, or geographical content reveals nuanced performance shifts over training steps. Task-specific evaluation metrics (e.g., factual accuracy, coherence, and instruction following) complement standard metrics like perplexity and cross-entropy loss.

In addition to subjective assessment, we further analyzed the relationship between the average qualitative score and three key variables: training steps, validation loss, and Hellaswag~\citep{zellers2019hellaswag} scores (see Table~\ref{tab:validation_hellaswag} for the progression of these metrics). Correlation analysis yielded the following results:

\begin{itemize}
    \item \textbf{Steps vs. Average Score:} $r = 0.928$, $R^2 = 0.860$.  
    As the number of training steps increases, the model’s qualitative performance improves substantially.

    \item \textbf{Validation Loss vs. Average Score:} $r = -0.945$, $R^2 = 0.893$.  
    The strong negative correlation indicates that decreasing validation loss closely aligns with improvements in the overall quality of generated text.

    \item \textbf{Hellaswag Score vs. Average Score:} $r = 0.963$, $R^2 = 0.927$.  
    Hellaswag scores correlate most strongly with qualitative improvements, suggesting that as the model becomes better at commonsense reasoning tasks, it also tends to produce higher-quality text across our evaluation criteria.
\end{itemize}

These findings confirm that while decreasing validation loss is a useful proxy for enhanced quality, complementary metrics like Hellaswag can provide even stronger predictive power. In other words, as the model’s performance on downstream benchmarks improves, so does its overall text quality as judged by human evaluators. The strong alignment between subjective judgments and objective metrics reinforces the notion that combining standard validation loss monitoring with targeted evaluations (such as Hellaswag) offers a robust framework for guiding and understanding the training process of large language models.

\begin{table}[!h]
\centering
\caption{Evaluation Scores for Different Training Steps}
\label{tab:evaluation_scores}
\begin{tabular}{c|cccccc|c}
\toprule
\textbf{Model Steps} & \textbf{C} & \textbf{F} & \textbf{R} & \textbf{FA} & \textbf{D} & \textbf{Cr} & \textbf{Average} \\
\midrule  
2700  & 4 & 4 & 5 & 2 & 3 & 3 & 3.50 \\
\hline
3300  & 5 & 5 & 6 & 3 & 4 & 4 & 4.50 \\
\hline
5100  & 5 & 5 & 6 & 3 & 4 & 4 & 4.50 \\
\hline
7500  & 6 & 6 & 6 & 3 & 4 & 5 & 5.00 \\
\hline
15700 & 6 & 6 & 7 & 4 & 5 & 5 & 5.50 \\
\hline
20000 & 6 & 6 & 7 & 4 & 5 & 5 & 5.50 \\
\hline
23900 & 7 & 6 & 7 & 4 & 6 & 6 & 6.00 \\
\hline
35000 & 7 & 7 & 7 & 4 & 6 & 6 & 6.17 \\
\hline
40500 & 7 & 7 & 7 & 5 & 6 & 6 & 6.33 \\
\hline
43250 & 7 & 7 & 8 & 5 & 7 & 7 & 6.83 \\
\bottomrule
\end{tabular}
\end{table}
\subsection{Comparison and Discussion}
\label{compdis}

The performance of DMaS-LLaMa-Lite is compared against TinyLLaMa's 2T checkpoint~\citep{zhang2024tinyllama}, a series of lightweight LLaMa-based models trained with significantly larger token counts. TinyLLaMa's 2T checkpoint, a 1.1B-parameter model trained on 2 trillion tokens using the SlimPajama dataset~\citep{cerebras2023slimpajama}, represents an established baseline for small-scale LLMs.

The benchmarks used for evaluation are as follows:

\begin{enumerate}
    \item \textbf{ARC Challenge}: A subset of the AI2 Reasoning Challenge \citep{clark2018think}, which focuses on grade-school science questions that require reasoning and commonsense knowledge.
    \item \textbf{ARC Easy}: The easier subset of ARC questions, designed to be straightforward but still requiring knowledge retrieval.
    \item \textbf{BoolQ}: A yes/no question-answering benchmark \citep{clark2019boolq} that requires reading comprehension and contextual reasoning.
    \item \textbf{HellaSwag}: A commonsense reasoning benchmark \citep{zellers2019hellaswag} where the goal is to select the most plausible continuation of a given sentence.
    \item \textbf{OpenBookQA}: A multiple-choice question benchmark \citep{mihaylov2018openbookqa} that requires reasoning over a small “open book” of facts.
    \item \textbf{PIQA}: The Physical Interaction QA benchmark \citep{bisk2020piqa}, which tests commonsense physical reasoning, such as how objects are used.
    \item \textbf{Winogrande}: A large-scale Winograd Schema Challenge \citep{sakaguchi2020winogrande} dataset, focusing on resolving pronoun ambiguity using commonsense reasoning.
\end{enumerate}

Figure~\ref{fig:dm_vs_tinyllama} illustrates the performance of DMaS-LLaMa-Lite (solid lines with dots) across various training steps alongside TinyLLaMa 2T (horizontal dotted lines). Each benchmark is color-coded for clarity. Key observations include:

\begin{figure}[htbp]
\centering
\includegraphics[width=15cm]{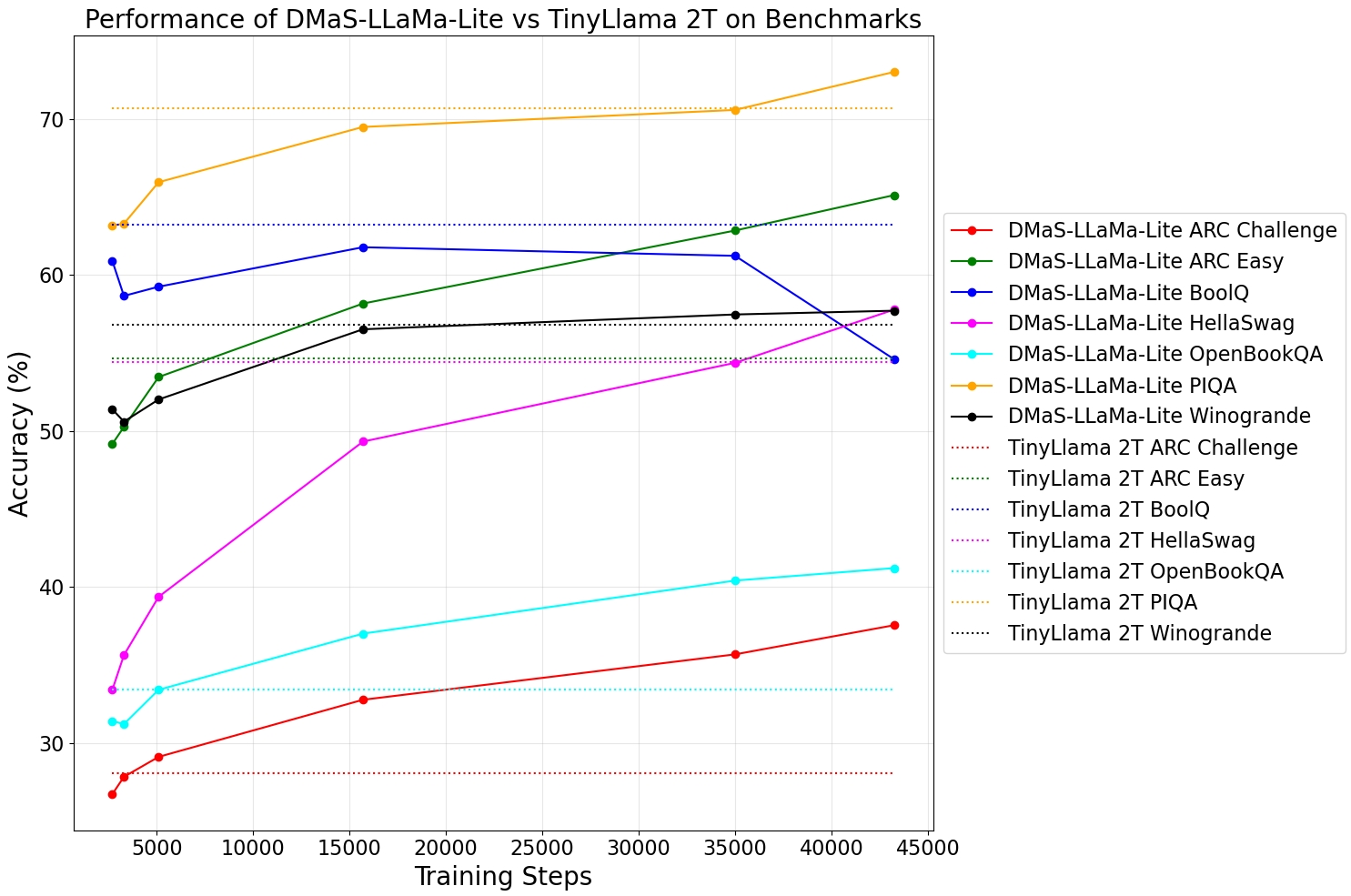} 
\caption{Performance comparison of DMaS-LLaMa-Lite checkpoints and TinyLLaMa (2T) across various benchmarks. Solid lines represent DMaS-LLaMa-Lite performance at different training steps, while horizontal dotted lines indicate TinyLLaMa 2T results.}
\label{fig:dm_vs_tinyllama}
\end{figure}

DMaS-LLaMa-Lite demonstrates superior performance on most benchmarks, such as ARC Challenge, ARC Easy, and OpenBookQA over the TinyLLaMa checkpoint, which was trained on 2 trillion tokens. We attribute these improvements to two key factors:

\begin{itemize}
    \item \textbf{Data Curation:} Our model was trained on the \texttt{HuggingFaceFW/fineweb-edu 100BT}\footnote{\url{https://huggingface.co/datasets/HuggingFaceFW/fineweb-edu}} dataset, a more carefully curated corpus compared to the SlimPajama\footnote{\url{https://huggingface.co/datasets/cerebras/SlimPajama-627B}} dataset used for TinyLLaMa. The higher quality of our training data likely facilitated more efficient learning, allowing the model to achieve strong performance on Hellaswag with far fewer tokens.
    \item \textbf{Model Scaling:} Although parameter count is not the sole determinant of downstream performance, increasing model capacity can enable more effective knowledge storage and representation. In this case, the jump from 1.1B to 1.7B parameters, combined with a better dataset, contributed to improved downstream task performance.
\end{itemize}

A notable exception to the general trend is the BoolQ benchmark~\citep{clark2019boolq}. Our model’s performance not only lags behind TinyLLaMa at 2T tokens but also diminishes slightly as training progresses. Intriguingly, this degradation in BoolQ performance parallels a similar pattern observed in TinyLLaMa: while TinyLLaMa initially achieves a BoolQ accuracy of around 63.21\% at 2T tokens, its accuracy declines to 57.83\% by the time it reaches 3T tokens of training. This phenomenon suggests a shared difficulty that may not stem from data quality or scale alone but from the intrinsic complexity of BoolQ-style yes/no question answering.

BoolQ questions often require more than pattern matching or simple retrieval of facts; they demand nuanced understanding and inference. For instance, consider the BoolQ example:

\begin{quote}
\textbf{Passage:} "Windows Movie Maker (formerly known as Windows Live Movie Maker in Windows 7) is a discontinued video editing software by Microsoft. It is a part of Windows Essentials software suite and offers the ability to create and edit videos as well as to publish them on OneDrive, Facebook, Vimeo, YouTube, and Flickr."

\textbf{Question:} "Is Windows Movie Maker part of Windows Essentials?"

\textbf{Answer:} "Yes"
\end{quote}

On the surface, identifying that Windows Movie Maker is part of Windows Essentials might seem straightforward. However, the presence of a potentially misleading term like “discontinued” can prompt the model—or even a human reader—to second-guess the relevance of the information to the current state of the software. This subtlety reveals that the model must not only locate relevant information but also correctly interpret it in context. Superficial pattern matching fails because the model has to navigate a small but meaningful inferential step: recognizing that being “discontinued” does not negate the fact that it was historically part of the Windows Essentials suite. 
    
This subtlety underscores that BoolQ tasks require interpretable reasoning. Yet, \textbf{the benchmark format itself does not give the model an opportunity to produce an explicit chain-of-thought or reasoning trace to use test time computation}~\citep{snell2024scaling}, but rather the task is framed to predict "yes" or "no" based on the passage content. Without such a mechanism, the model must supply an immediate yes/no answer, leaving no room to display intermediate reasoning steps that might help ensure correctness. Thus, the challenge is twofold: models need robust reasoning capabilities, and the benchmark structure (which does not request or evaluate reasoning traces) limits the model’s ability to leverage explicit reasoning strategies that could improve performance.

As a result, smaller models—trained only through standard next-token prediction—may not naturally acquire the level of interpretative reasoning necessary to excel at tasks like BoolQ. They have neither the inherent capability nor the format in the benchmark to showcase a reasoning process. This shortcoming suggests that more specialized training (e.g., instruction tuning) or evaluation formats that encourage reasoning traces might be required to handle nuanced yes/no inference tasks effectively.

\begin{table}[htbp]
\centering
\caption{Validation Loss and Hellaswag Scores Across Training Steps}
\label{tab:validation_hellaswag}
\begin{tabular}{c|c|c}
\toprule
\textbf{Steps} & \textbf{Validation Loss} & \textbf{Hellaswag Score} \\
\midrule 
2700 & 3.00 & 31.9 \\\hline
3300 & 2.93 & 32.9 \\\hline
5100 & 2.79 & 37.7 \\\hline
7500 & 2.71 & 40.5 \\\hline
15700 & 2.65 & 46.9 \\\hline
20000 & 2.55 & 49.6 \\\hline
23900 & 2.50 & 50.2 \\\hline
35000 & 2.48 & 52.5 \\\hline
40500 & 2.46 & 53.2 \\\hline
43250 & 2.47 & 56.1 \\
\bottomrule
\end{tabular}
\end{table}

\section{Post-training}
\label{posttrain}

Following the initial pre-training phase, we conducted a post-training step to refine the model’s capacity for following user instructions and producing contextually appropriate, helpful responses. This stage leverages techniques akin to instruction tuning, where a base model—already competent in general text generation—is further specialized to respond more effectively to prompts in a conversational format.

\subsection{Instruction-Following Data and Template}

For the post-training stage, we used the \texttt{yahma/alpaca-cleaned}~\footnote{\url{https://huggingface.co/datasets/yahma/alpaca-cleaned}} dataset, a widely utilized instruction-tuning corpus derived from the Stanford Alpaca dataset~\citep{alpaca}. To present this data to the model in a consistent, conversational manner, we applied a Vicuna 1.1-style chat template~\citep{zheng2023judging}, which structures the interaction as a user-assistant conversation:

\texttt{A chat between a curious user and an artificial intelligence assistant. The assistant}\\\texttt{ gives helpful, detailed, and polite answers to the user's questions.}

\texttt{USER: {input}}

\texttt{ASSISTANT: {output}<|endoftext|>} 

Here, \texttt{{input}} is replaced with the user's query or instruction, and \texttt{{output}} with the model’s expected answer. The <|endoftext|> token serves as a signal for the model to stop generating text at the appropriate point.

This template encourages the model to produce answers in a polite, contextually aligned, and helpful manner, while also teaching it to respect conversational boundaries. Since the training data already includes well-structured instructions and expert-crafted responses, the model can internalize the patterns and nuances of effective assistant behavior.

\subsection{Model Initialization and Parameter-Efficient Fine-Tuning}

We began the post-training step from the 43,500-step checkpoint of DMaS-LLaMa-Lite. Rather than fine-tuning all parameters, we employed a parameter-efficient technique based on Low-Rank Adaptation (LoRA)~\citep{hu2021lora}. We injected LoRA adapters into attention-related parameters (e.g., \texttt{q\_proj}, \texttt{k\_proj}, \texttt{v\_proj}, \texttt{o\_proj}, and feed-forward projections) using a rank of $r=128$, a LoRA $\alpha=16$, and no dropout. This approach reduces the memory footprint and computational overhead while enabling efficient adaptation of the model to the instruction-following domain.

The model’s training precision and device configuration were automatically adjusted based on hardware capabilities. In this case, half-precision (FP16) or bfloat16 was used where available. Batch sizes, gradient accumulation steps, and other hyperparameters were chosen to balance computational feasibility with adequate training signal. Specifically, we set a per-device batch size of 32 and a gradient accumulation of 16 to maintain stable gradient updates, and we employed the AdamW 8-bit optimizer for memory-efficient training.

\subsection{Training Dynamics}

\begin{table}[!h]
\centering
\caption{Comparison of Responses Before and After Post-Training Instruction Tuning}
\label{tab:qual_comparisons}
\begin{tabular}{p{0.15\textwidth} | p{0.35\textwidth} | p{0.35\textwidth}}
\toprule
\textbf{Prompt} & \textbf{Before Post-Training} & \textbf{After Post-Training} \\
\midrule
\begin{tabular}[c]{@{}l@{}}\\ \\Who invented \\ the first \\ practical light \\ bulb?\end{tabular} & 
\begin{quote}Thomas Edison.\end{quote} & 
\begin{quote}The first practical light bulb was invented by Thomas Edison in 1879. He was a prolific inventor and held over 1,000 patents in his name...\end{quote} \\

\hline

\begin{tabular}[c]{@{}l@{}}\\ \\ What are the \\ Pyramids of Giza \\ known for?\end{tabular} & 
\begin{quote}They are known for being the largest structures ever built by man.\end{quote} &
\begin{quote}The Pyramids of Giza are known for their architectural and engineering brilliance. They were built by the ancient Egyptians as burial monuments for their pharaohs. The pyramids are considered to be the most famous and largest of the three pyramids in the Giza...\end{quote} \\

\bottomrule
\end{tabular}
\end{table}
The post-training step encompassed approximately 400 steps of fine-tuning, with a linear learning rate schedule peaking at $5 \times 10^{-5}$. We applied a brief warmup period of five steps to mitigate instability at the onset. During this process, the training loss steadily decreased from approximately 2.25 at the start of instruction tuning to about 1.20 by the end of the 400 steps, reflecting substantial gains in the model’s ability to generate coherent, contextually appropriate answers aligned with user instructions.

It is worth noting that the fine-tuning process does not rely on massive training resources or prolonged runs; instead, the combination of LoRA adapters and a high-quality instruction dataset enables rapid improvements. These modifications help the model internalize the format, style, and content typical of user-posed questions and the expected assistant-style responses, ultimately guiding the model toward more useful and context-sensitive outputs.

\subsection{Qualitative Improvements in Responses}

Table~\ref{tab:qual_comparisons} compares the conversational behaviors of the models before and after post-training using two example prompts. Interestingly, the model demonstrates an ability to answer questions even prior to post-training. This phenomenon can be attributed to two factors. First, the Vicuna 1.1 prompting format aligns closely with natural conversational scenarios and clearly indicates that the assistant is expected to provide answers. Consequently, the model can generate responses based on its pre-training alone, without requiring additional fine-tuning. Second, it is possible that the pre-training dataset includes instruction-tuning samples formatted similarly to Vicuna 1.1 prompts, as this format predates the pre-training corpus's knowledge cutoff.

Another noteworthy observation is that instruction tuning during post-training significantly enhances the detail and contextual depth of the model's responses. Compared to the pre-training-only model, the post-trained version consistently generates more elaborate and contextually rich answers, reflecting improved understanding and engagement with the prompts.

\subsection{Sensitivity to Prompt Variations}

An intriguing phenomenon we observed is the model’s sensitivity to a small change in the prompt formatting. For example, simply appending a space after “ASSISTANT:” can alter the generated response. Without the space, the model might produce a normal answer. With the space, it often continues with a more elaborate completion:

\texttt{ASSISTANT: \_\_\_\_\_\_\_\_\_\_\_\_\_\_\_\_\_\_\_\_\_\_\_\_\_\_\_\_\_}

\texttt{The first practical light bulb was invented by Thomas Edison in 1879...} 

One plausible explanation involves the underlying tokenizer and how it handles spaces. In GPT-2-style tokenizers, spaces are generally prefixed to words rather than suffixed. This leads to subtle differences in the tokenization of seemingly similar inputs. For example, for an answer in the training set: "ASSISTANT: Berlin was founded in the 13th century.", it is tokenized as: ["ASS", "IST", "ANT", ":", "ĠBerlin", "Ġwas", "Ġfounded", "Ġin", "Ġthe", "Ġ13", "th", "Ġcentury", "."]. Here, “Ġ” represents a space token. If the prompt ends with "ASSISTANT: ", however, the tokenization changes to: ["ASS", "IST", "ANT", ":", "Ġ"]. The inclusion of the extra "Ġ" token alters the model’s understanding of the input and significantly shifts the probability distribution for the next token. This is due to the statistical co-occurrence patterns learned during pre-training, which differ depending on the presence of the additional space token.

This observation underscores the model’s inherent sensitivity to input formatting and highlights the critical role of precise prompt engineering, even down to seemingly trivial details like whitespace handling.

\section{Conclusion}
\label{conclu}

In this paper, we documented the pre-training experience of DMaS-LLaMa-Lite, a fully open-source, 1.7-billion-parameter LLaMa-based language model trained on approximately 20 billion tokens of a carefully curated dataset. By chronicling the model’s training trajectory through validation loss, downstream benchmarks, qualitative evaluations, and practical challenges, we demonstrated how the training process influences model outputs and performance.

Our analysis showed that model quality and coherence improve steadily as validation loss decreases. Early checkpoints produce semantically incoherent or factually dubious content, but with more training steps, completions become increasingly fluent, contextually relevant, and factually accurate. Correlation analyses further indicate that improvements in validation loss and downstream scores (e.g., Hellaswag) closely mirror enhancements in subjective quality measures.

The training journey underscored key best practices and pitfalls. We found that maintaining continuity in training is paramount; resuming training without restoring optimizer states led to abrupt increases in validation loss and required substantial additional steps for recovery. Transitioning between different hardware configurations also impacted stability and throughput, suggesting that consistent training environments and procedures can enhance model performance and convergence speed.

Post-training instruction tuning further refined the model's capabilities, enabling it to respond more elaborately to a wide variety of prompts. By using parameter-efficient fine-tuning with high-quality instruction datasets, we observed significant behavioral shifts in the model. And we observed the sensitivity of the model to prompting details. 

Importantly, we demonstrated that fewer tokens of high-quality data can outperform vast but less curated corpora, as evidenced by comparisons against models like TinyLLaMa. By leveraging a carefully filtered dataset, our model achieved competitive downstream performance with significantly fewer training tokens. However, some persistent challenges—such as reduced performance on BoolQ and limited factual accuracy early in training—highlight that certain tasks demand more complex reasoning capabilities. For these tasks, additional instruction tuning or training techniques that elicit reasoning traces may be necessary.

Our findings, along with the provided training logs, checkpoints, and sample outputs, aim to guide future researchers and practitioners in refining both pre-training and post-training strategies. By documenting the full trajectory from pre-training dynamics to post-training refinement, we emphasize a process-focused perspective where transparency and detailed insights foster reproducibility and more effective model development. This approach lays a foundation for future studies to explore additional best practices, methodological refinements, and novel techniques in both pre-training and instruction-tuning phases of large language models.

\section*{Acknowledgements}

The research is supported in part by Discovery Grants (RGPIN-2024-04087) from the Natural Sciences and Engineering Research Council of Canada (NSERC), Canada Research Chairs Program (CRC-2019-00041), and R\&D Spearhead Grants (2024-1301) from the National Cybersecurity Consortium (NCC).

\bibliographystyle{cas-model2-names}
\bibliography{main}

\bio{}
\textbf{Miles Q. Li} is a postdoctoral researcher in the School of Information Studies at McGill University, Montreal, Canada. He received his B.Sc and M.Sc from Peking University and his Ph.D. from McGill University. His research interests include deep learning and its application in natural language processing and cybersecurity.
\endbio

\bio{} 
\textbf{Benjamin C. M. Fung} received his Ph.D. degree in Computing Science from Simon Fraser University in Canada in 2007. He is a Canada Research Chair in Data Mining for Cybersecurity and a Professor at the School of Information Studies, McGill University, Canada. He also serves as Associate Editor for IEEE Transactions of Knowledge and Data Engineering (TKDE) and Elsevier Sustainable Cities and Society (SCS). He has over 150 refereed publications, with more than 16,000 citations, that span the research forums of data mining, privacy protection, cybersecurity, services computing, and building engineering. Prof. Fung is also a licensed Professional Engineer of software engineering in Ontario, Canada.
\endbio

\bio{} 
\textbf{Shih-Chia Huang} is a Full Professor in the Department of Electronic Engineering at National Taipei University of Technology, Taiwan, and an International Adjunct Professor with the Faculty of Business and Information Technology at the University of Ontario Institute of Technology, Canada. He has been named a senior member of the Institute of Electrical and Electronic Engineers (IEEE). He
is currently the Chair of the IEEE Taipei Section Broadcast Technology Society, and was a Review Panel Member of the Small Business Innovation Research (SBIR) program for the Department of Economic Development of Taipei City and New Taipei City, respectively. His research interests include intelligent multimedia systems, image processing and video coding, video surveillance systems, cloud computing and big data analytics, artificial intelligence, and mobile applications and systems.

\end{document}